\documentclass{article}

     \PassOptionsToPackage{numbers, compress}{natbib}

\usepackage{multirow}
\usepackage{neurips_2022}
\usepackage{svg}

\usepackage[utf8]{inputenc} 
\usepackage[T1]{fontenc}    
\usepackage{hyperref}       
\usepackage{url}            
\usepackage{booktabs}       
\usepackage{amsfonts}       
\usepackage{nicefrac}       
\usepackage{microtype}      
\usepackage{xcolor}         
\usepackage{graphicx}
\usepackage{amsmath}
\usepackage{setspace}

\title{Dissimilar Nodes Improve Graph Active Learning}

\author{Zhicheng Ren \\
University of California, Los Angeles \\
\texttt{franklinnwren@g.ucla.edu} \\
\And
Yifu Yuan \\
University of California, Los Angeles \\
\texttt{yiv.yuanyifu@hotmail.com} \\
\And
Yuxin Wu \\
University of California, Los Angeles \\
\texttt{yuxinwu98611@g.ucla.edu} \\
\And
Xiaxuan Gao \\
University of California, Los Angeles \\
\texttt{gaox8499@gmail.com} \\
\And
Yewen Wang \\
University of California, Los Angeles \\
\texttt{wyw10804@gmail.com} \\
\And
Yizhou Sun \\
University of California, Los Angeles \\
\texttt{yzsun@cs.ucla.edu} \\
}

\begin{document}

\maketitle

\begin{abstract}

Training labels for graph embedding algorithms could be costly to obtain in many practical scenarios. Active learning (AL) algorithms are very helpful to obtain the most useful labels for training while keeping the total number of label queries under a certain budget. The existing Active Graph Embedding framework proposes to use centrality score, density score, and entropy score to evaluate the value of unlabeled nodes, and it has been shown to be capable of bringing some improvement to the node classification tasks of Graph Convolutional Networks. However, when evaluating the importance of unlabeled nodes, it fails to consider the influence of existing labeled nodes on the value of unlabeled nodes. In other words, given the same unlabeled node, the computed informative score is always the same and is agnostic to the labeled node set. With the aim to address this limitation, in this work, we introduce 3 dissimilarity-based information scores for active learning: feature dissimilarity score (FDS), structure dissimilarity score (SDS), and embedding dissimilarity score (EDS). We find out that those three scores are able to take the influence of the labeled set on the value of unlabeled candidates into consideration, boosting our AL performance. According to experiments, our newly proposed scores boost the classification accuracy by 2.1$\%$ on average and are capable of generalizing to different Graph Neural Network architectures.
\end{abstract}

\section{Introduction}
Graphs are a great data type that can perfectly model entities and their interactions. Therefore, they are everywhere in our daily life, such as social networks and natural science. Nowadays, Graph Neural Networks (GNNs) have been drawing increasing research attention because of their great success in various applications \cite{kipf2017semi} \cite{10.5555/3294771.3294869} \cite{velickovic2018graph} \cite{schlichtkrull2018modeling}\cite{xu2018how} \cite{10.5555/3327345.3327389}. However, in many real-world scenarios, such as chemistry and health care, it could be very expensive to collect a sufficient amount of labeled data to facilitate the training of GNNs, which limits the performance of GNNs.

Active learning is a framework to resolve this challenge, which allows dynamic query of new node labels from unlabeled node sets given a limited budget, and has been shown to be very helpful in various learning problems \cite{beluch2018power} \cite{gal2017deep} \cite{10.1007/978-3-319-10593-2_37}. AL algorithm generally consists of two main components: a query system to select the most valuable instances for the downstream task, and an oracle to label the selected sample. In recent years, several frameworks like \cite{DBLP:journals/corr/CaiZC17} \cite{ma2022partition} \cite{wu2019}
proposes AL algorithms for graphs in order to handle the unique challenge when applying active learning to graph data: graph data is not independent and identically distributed, which requests us to consider how to incorporate graph structure into the query system design to capture the correlation among different nodes.

However, existing graph active learning query functions only use properties of the unlabeled nodes to determine the next label to be queried \cite{DBLP:journals/corr/CaiZC17}. In many real-world data sets, however, we often do not have control over the distribution of the available labeled set of nodes. The available labeled node set could be biased (e.g. most of them belong to a single class) or isolated (e.g. all of them are in a single isolated sub-graph). If we need to query a new label, the label which is not as valuable given an unbiased labeled set could turn out to be very valuable given a biased labeled set. Hence, designing label query functions that consider the influence of the current labeled set on the value of candidates for label queries will be promising in selecting the most valuable labels. 

In this work, inspired by the idea in \cite{10.1145/3472291} that a more diverse training set improves the performance of active learning, we propose three novel active learning scores for graphs based on node dissimilarity: feature dissimilarity score (FDS), structure dissimilarity score (SDS), and embedding dissimilarity score (EDS). Those scores directly evaluate how dissimilar a candidate node for label queries is with respect to existing labeled nodes. We conduct extensive experiments to demonstrate that those scores improve the performance of node classification tasks of GCN by about 2.1$\%$ when added to the other conventionally-used active learning scores. Meanwhile, we also conduct some ablation studies where we replace GCN backbone with other well-known graph neural network variants such as GAT \cite{velickovic2018graph} and SGC \cite{pmlr-v97-wu19e}. Results show that our methods are generalizable to those GNN variants.

\section{Background and Related Works}

\subsection{Graph Neural Network}
GNN is a family of graph embedding models that achieves state-of-the-art performance in graph-related tasks. It learns node representations by iteratively aggregating neighborhood information using a convolution operator, and most of the GNN architecture designs present a GNN layer in the following message-passing form:

\begin{equation}
    \textbf{H}^{(l)} = \sigma(\Tilde{\textbf{A}}\textbf{H}^{(l-1)}\textbf{W}^{(l)})
\end{equation}

Here, $\textbf{H}^{(l)}$ is the hidden node representation at layer $l$, $\sigma$ is the non-linear activation, $\Tilde{\textbf{A}}$ is the graph convolutional filter, and $\textbf{W}^{(l)}$ is the learnable weight at layer $l$.

Various GNN architectures are proposed during the past years. GCN \cite{kipf2017semi}  takes the first step to formally propose the graph convolutional operation. GAT \cite{velickovic2018graph} improves the model architecture with the attention mechanism. GraphSAGE \cite{10.5555/3294771.3294869} proposes to use a sampled neighborhood for message passing, which improves the efficiency of GCN and enables it to be applicable in an inductive setting. SGC \cite{pmlr-v97-wu19e} finds that after removing nonlinearities and collapsing weight matrices between consecutive layers,  GCN would not suffer from performance drop, and can get better scalability. GCNII \cite{chen2020simple} uses the techniques of initial residual and identity mapping to alleviate the over-smoothing problem of GCN. MixHop \cite{abu2019mixhop} concatenates the aggregated embeddings obtained by using the k-hop random walk transition matrices for different k, which enables the model to get information from indirect neighbors in one layer.

\subsection{Active Learning}
Active learning (AL) is a framework that selects the most valuable samples from the unlabeled set under the given budget and queries their labels, in order to improve the performance of downstream tasks. AL consists a query system and an oracle. The query system selects the most valuable instances from the unlabeled dataset, and the oracle then annotates the selected instances. Most of the query systems can be categorized into four types \cite{10.1145/3472291}: (1) uncertainty-based query systems select the samples according to their uncertainty ranking, and tend to pick the most uncertain sample; (2) diversity-based query systems that select samples sharing the most diverse attributes; (3) performance-based query systems which measure the expected effect of selected samples on the model training; (4) representativeness-based query systems which select samples that could best represent the true data distribution.

\subsection{Notable Graph Active Learning Frameworks}

In recent years, as GNNs become increasingly popular, graph active learning gets more research attention. Here are some examples of graph active learning works from different AL categories:

\textbf{Uncertainty-based query systems.} The work of \cite{DBLP:journals/corr/CaiZC17} proposes a strong graph active learning baseline which combines uncertainty-based query systems and representativeness-based query systems. This work shows that uncertainty-based query scores need to be combined with other metrics to select the most valuable nodes at the early training stage, since the model could not provide sufficient probability information at the early training stage.

\textbf{Representativeness-based query systems.} Apart from \cite{DBLP:journals/corr/CaiZC17}, there are some other graph active learning frameworks that use representativeness-based query systems. Wu et. al. \cite{wu2019} improve on the KMeans clustering method and yield better accuracy among representativeness-based metrics. Recently, Ma et. al. \cite{ma2022partition} propose a graph partition algorithm to enhance the representativeness of the selected nodes. It is notable that both works focus on improving representativeness-based metrics only, which might neglect certain under-represented nodes.

\textbf{Performance-based query systems.} Hu et. al. \cite{10.5555/3495724.3496577} propose a performance-based active learning framework based on reinforcement learning.
Although this work thoroughly exploits the strength of performance-based metrics, the reinforcement learning framework itself could be a significant computational burden.

Representativeness-based metrics might fail to address the nodes which have low centrality, while performance-based metrics need pre-computation to evaluate the value of a new label, causing a significant computational burden. On the other hand, uncertainty-based needs to be combined with other metrics to select the most valuable nodes at the early training stage since the model could not provide sufficient probability information at the early training stage. Therefore, we decide to explore diversity-based query systems. We find out that, by directly comparing the labeled node set and the unlabeled node set, we are able to involve the influence of the labeled set on the value of unlabeled candidates into consideration, boosting our AL performance.

\section{Method}

\subsection{Workflow of Graph Active Learning}
\begin{figure}[htbp]
\centering
\includegraphics[width=10cm]{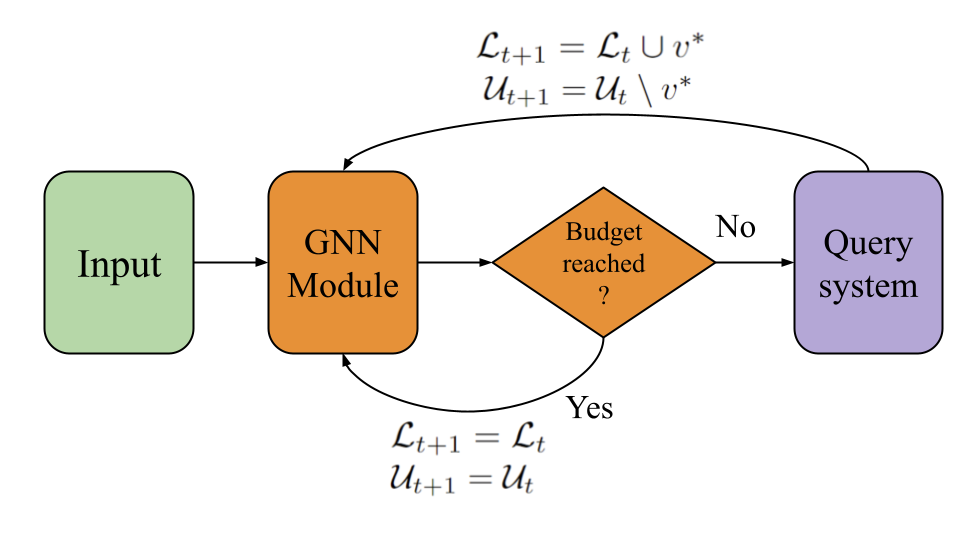}
\caption{Graph Active Learning Framework. $v^*$ is the node being selected at epoch $t$. ${\mathcal{L}_t}$, ${\mathcal{U}_t}$} are labeled and unlabeled node sets of our input graph at time $t$ respectively.
\end{figure}
Our work focus on the node classification task. Figure 1 illustrates a general framework of our methodology. Given an attributed graph $\mathcal{G} = (\mathcal{V},\mathcal{E})$ with node set $\mathcal{V}$ and edge set $\mathcal{E}$, $\mathcal{V}$ would be the union set of the labeled node set ${\mathcal{L}_0}$ and unlabeled node set ${\mathcal{U}_0}$ at epoch 0. Now, we let the node attribute matrix be $X$, the class vector for $\mathcal{V}$ be $Y$, the labeling budget for AL be $B$, and the total training epoch be $T$.

We follow the training process of the standard active learning framework. At every training epoch $t$, if the labeling budget $B$ is not reached, the AL query strategy module selects the best candidate in the unlabeled nodes set at the current epoch (denoted as $\mathcal{U}_t$) with the query strategy, queries its label with the oracle, and adds it into the labeled nodes set at current epoch (denoted as $\mathcal{L}_t$), i.e. we will have $\mathcal{U}_{t+1} = \mathcal{U}_t \setminus v^*$ and $\mathcal{L}_{t+1} = \mathcal{L}_t \cup v^*$ for epoch $t+1$ where $v^*$ is the node being selected at epoch $t$. If the labeling budget $B$ is reached, i.e., $| \mathcal{L}_t | = B$, we stop the "selecting and querying" process. The training is carried out normally until GNN converges. Let $l$ be the loss of the GNN model training, the optimization objective of the graph active learning problem could be formally formulated as:

\begin{equation}
    \underset{\mathcal{L}_T, \mathcal{U}_T}{\text{argmin}} \ \mathbb{E} \left[ \ell (\text{GNN}(\mathcal{G}, X), Y) \right], \ \text{such that} \ |\mathcal{L}_T| = B
\end{equation}

\subsection{AGE Scores}

Active Graph Embedding (AGE) is a framework proposed by \cite{DBLP:journals/corr/CaiZC17}. AGE uses $\phi_{\text{AGE}}$ as a query function to select the best querying candidates. $\phi_{\text{AGE}}$ has three components.

The first and the second component of $\phi_{\text{AGE}}$ are two representativeness-based metrics: graph centrality and information density.  Graph centrality measures how important a candidate node $v_i$ is in the perspective of information flow on the entire graph by adopting the PageRank centrality \cite{ilprints422}. Information density measures how representative a candidate node $v_i$ is by using KMeans clustering, the closer a node is to the cluster center, the more representative it is. Their equations are as follows:
\begin{equation}
    \phi_{\text {centrality }}\left(v_{i}\right)=\rho \sum_{j} A_{i j} \frac{\phi_{\text{centrality}}\left(v_{j}\right)}{\sum_{k} A_{j k}}+\frac{1-\rho}{N}
\end{equation}
\begin{equation}
    \phi_{\text {density }}\left(v_{i}\right)=\frac{1}{1 + \text{ED} \left(\text{EMB}(v_{i}), \text{CC}(v_{i})\right)}
\end{equation}

where $A$ is the adjacency matrix, ED is the euclidean distance function, $\text{EMB}(v_i)$ is the embedding of the node $v_i$, $\text{CC}(v_i)$ is the center of the cluster which $v_i$ belongs to, and $\rho$ is the damping parameter.

The third one is an uncertainty-based metric that uses the entropy information of each node during the learning process. It is calculated according to the following equation:
\begin{equation}
    \phi_{\text{entropy}}\left(v_{i}\right)=-\sum_{c=1}^{C} \mathbf{P}\left(Y_{i c}=1 \mid \mathcal{G}, \mathcal{L}, X\right) \log \mathbf{P}\left(Y_{i c}=1 \mid \mathcal{G}, \mathcal{L}, X\right),
\end{equation}
\\
where the component $\mathbf{P}\left(Y_{i c}=1 \mid \mathcal{G}, \mathcal{L}, X\right)$ denotes the probability that node $v_i$ belongs to class $c$. 

Define $r_{\phi_{*}}(v_i)$ as the high-to-low ranking percentage of score $\phi_*$ of example $v_i$. Then, the AGE score of example $v_i$ is:
\begin{align}
    \phi_{\text{AGE}}\left(v_{i}\right) &= \alpha_1 r_{\phi_{\text {centrality }}}\left(v_{i}\right) + \alpha_2 r_{\phi_{\text {density }}}\left(v_{i}\right)
    + \alpha_3 r_{\phi_{\text{entropy}}}\left(v_{i}\right)
\end{align}

where $\alpha_1$, $\alpha_2$ and $\alpha_3$ are the coefficients of $\phi_{\text {centrality}}$, $\phi_{\text {density}}$ and $\phi_{\text{entropy}}$ in the AGE model.

\subsection{Dissimilarity Scores}

Inspired by the well-recognized idea in active learning that a more diverse training set can benefit the performance \cite{10.1145/3472291}, we propose three node dissimilarity-based metrics to augment the query strategy and improve the active graph embedding system.

\textbf{Feature dissimilarity score (FDS):} To evaluate how dissimilar a candidate node is compared to the nodes that have already been labeled in terms of their node feature, we propose FDS, which applies the commonly used cosine similarity for feature distance measurement. Let $\text{FEA}(v_{i})$ be the feature vector of node $v_{i}$, then:
\begin{equation}
    \phi_\text{FDS} \left(v_{i}\right) = \frac{1}{\underset{l \in \mathcal{L}_{t}}{\text{max}} \ \text{cos}(\text{FEA}(v_{i}), \text{FEA}(l))}
\end{equation}

\textbf{Structural dissimilarity score (SDS):} In the context of graph data sets, the similarity of two nodes is not only dependent on how similar their features are, but also dependent on their graph-level structures. To evaluate the structural dissimilarity, we consider the second-order similarity \cite{tang2015line}, with the intuition that whose underlying idea is, two nodes are similar if their neighbors are similar. Given the adjacency matrix $A$, the element $(A^2)_{i, j}$ gives the number of paths of length 2 from node $i$ to node $j$, i.e. number of shared neighbors between node $i$ and node $j$. Therefore, we could design our score $\phi_\text{SDS}$ to be:
\begin{equation}
    \phi_\text{SDS} \left(v_{i}\right) = \frac{1}{\underset{{l \in \mathcal{L}_t}}{\text{max}} \ (A^2)_{i, \text{index}(l)}}
\end{equation}
where $\text{index}(l)$ is the index of the labeled node $l$. We can interpret this idea as: if we already have a labeled node to propagate the feature information to its neighbors, the value of another node label connected to the same group of neighbors might be low in the setting of GNNs.

\textbf{Embedding dissimilarity score (EDS):} EDS is designed to measure node dissimilarity based on their embeddings. It utilizes the node representations obtained by the GNN backbone model, and it is capable of capturing both structure and feature information. Similar to FDS, we use cosine similarity as a distance measurement.
\begin{equation}
    \phi_\text{EDS} \left(v_{i}\right) = \frac{1}{\underset{{l \in \mathcal{L}_t}}{\text{max}} \ \text{cos}(\text{EMB}(v_{i}), \text{EMB}(l))}
\end{equation}
where EMB($l$) is the embedding of node $l$ after the convolution.

The main difference between our dissimilarity-based metrics and previous graph active learning metrics is that by directly comparing between $\mathcal{L}_t$ and $\mathcal{U}_t$, we are able to take the influence of the labeled set on the value of unlabeled candidates into consideration. 

Our aggregated $\phi_{\text{dissimilarity}}$ is formulated in this way:
\begin{align}
    \phi_{\text{dissimilarity}}\left(v_{i}\right) &= \beta_1 r_{\phi_{\text {FDS }}}\left(v_{i}\right) + \beta_2 r_{\phi_{\text {SDS }}}\left(v_{i}\right)
    + \beta_3 r_{\phi_{\text{EDS}}}\left(v_{i}\right)
\end{align}
Here, $\beta_1$, $\beta_2$, and $\beta_3$ are the coefficients of our proposed FDS, SDS, and EDS.

\subsection{Query Module}

In the fore-mentioned AL system, the most important part is the query strategy, which evaluates the importance of each unlabeled node. Combining our newly proposed dissimilarity scores in 3.3 and the conventional active graph embedding metrics \cite{DBLP:journals/corr/CaiZC17} in 3.2, our AL query module can be described as follows:
\begin{align*}
    v^* &= \underset{i}{\text{argmax}} \ (\alpha \phi_\text{AGE}\left(v_{i}\right) + \beta \phi_\text{dissimilarity}\left(v_{i}\right))
\end{align*}
where $\phi_\text{AGE}\left(v_{i}\right)$ is the traditional AGE query metrics mentioned in equation 6 and $\phi_\text{dissimilarity}\left(v_{i}\right)$ is the combination of the three newly proposed dissimilarity scores in equation 10. The hyper-parameter $\alpha$ and $\beta$ are time-sensitive. We learn our hyper-parameters with the following constraints:
\begin{align}
    \alpha + \beta &= 1\\
    \alpha_1 + \alpha_2 + \alpha_3 &= \alpha\\
    \beta_1 + \beta_2 + \beta_3 &= \beta
\end{align}

Inspired by \cite{DBLP:journals/corr/CaiZC17}, We learn our hyper-parameters in the following:
\begin{align}
    \alpha_1 &\sim \mathbf{Beta}(1, \ 1.005 - Ct)\\
    \alpha_2 &= \alpha_3 = \beta_1 = \beta_2 = \beta_3 = \frac{1 - \alpha_1}{5}
\end{align}
$t$ is the epoch number, $\mathbf{Beta}$ is the beta distribution where we sample $\alpha_1$ from, $C$ is a constant and a hyper-parameter to be chosen. On a high level, the time-sensitive hyper-parameter set increases the weight of $\phi_\text{centrality}\left(v_{i}\right)$ at the early training stage because we want to collect more representative nodes from the graph, and decreases the weight of $\phi_\text{centrality}\left(v_{i}\right)$ at the later training stage since we want to include more dissimilar nodes to our existing labeled set. When running experiments on one of our dissimilarity scores (e.g. FDS), we let coefficients of other dissimilarity scores (e.g. $\beta_2$ and $\beta_3$ in the case of FDS) to be 0.

\section{Experiments and Discussion}

\subsection{Experimental Setting}
We evaluate the performance of our proposed scores on node classification tasks. We select three commonly used datasets: Cora \cite{mccallum2000automating}, Citeseer \cite{giles1998citeseer}, and Pubmed \cite{Sen_Namata_Bilgic_Getoor_Galligher_Eliassi-Rad_2008}. We select GCN \cite{kipf2017semi} as our GNN backbone, and compare it with the AGE-only \cite{DBLP:journals/corr/CaiZC17} , FeatProp (FP) \cite{wu2019}, GraphPart (GP) \cite{ma2022partition} and GraphPartFar (GPF) \cite{ma2022partition} baselines to show that our proposed scores can improve this promising system. We use Macro F1 and Micro F1 to evaluate our node classification performance. Details about the data set and our training settings are in Appendix A.

We did not compare with \cite{10.5555/3495724.3496577}, since it is a reinforcement learning framework that requires training on one dataset and testing on another dataset. This is different from our node classification task setting, where we try to predict the identity of the unlabeled nodes in the same dataset.

\subsection{Main Results}
We present our main results in Table 1. The highest F1 scores are in bold form, and the second-highest F1 scores are underlined. The baselines are on the left-hand side and our methods are on the right-hand side.

\begin{table*}[h!]
\centering
\scriptsize
\begin{tabular}{|c|c|ccccc|cccc|} 
 \hline
 \textbf{\tiny{Dataset}} & 
 \textbf{\tiny{Metric}} &\textbf{\tiny{GCN}} & \textbf{\tiny{AGE-only}} & \textbf{\tiny{FP}} & \textbf{\tiny{GP}} & \textbf{\tiny{GPF}} & \textbf{\tiny{AGE+FDS}} & \textbf{\tiny{AGE+SDS}} & \textbf{\tiny{AGE+EDS}} & \textbf{\tiny{AGE+FDS+SDS}}\\
 \hline
 
 \multirow{2}{*}{\textbf{\tiny{Cora}}} &   \textbf{\tiny{MacroF1}(\%)}  & 79.13 & 80.22 & 80.91 & 81.23 & \textbf{82.46} & \underline{81.54} & 81.00 & 80.85 & 81.43\\
 
&\textbf{\tiny{MicroF1}(\%)} & 79.95 & 81.36 & 82.22 & 82.41 & \textbf{83.41} & 82.55 & 82.18 & 81.84 & \underline{82.59}\\
  
\hline

 \multirow{2}{*}{\textbf{\tiny{Citeseer}}} &   \textbf{\tiny{MacroF1}(\%)} & 66.96 & 66.70 & 65.17 & 65.12 & 64.74 &  66.93 & \textbf{67.86} & 66.94 & \underline{67.43}\\

&\textbf{\tiny{MicroF1}(\%)} & 70.83 & 71.92 & 70.03 & 69.31 & 69.62 & \underline{72.77} & 72.69 & 72.42 & \textbf{73.06}\\

\hline
 \multirow{2}{*}{\textbf{\tiny{Pubmed}}} &   \textbf{\tiny{MacroF1}(\%)} & 77.30 & 78.85 & 75.65 & 78.30 & 77.29 & 79.37 & \underline{79.43} & 76.87 & \textbf{79.53}\\

&   \textbf{\tiny{MicroF1}(\%)} & 77.91 & 79.43 & 76.57 & 79.72 & 77.86 & \underline{80.00} & 79.93 & 77.90 & \textbf{80.23}\\

\hline
\end{tabular}
\caption{F1 Scores of Different Active Learning Score Combination for GCN Backbone}
\label{table:GCN}
\end{table*}

Compared to GCN and AGE-only, all three scores (FDS, SDS, and EDS) have shown improvement in the node classification accuracy on our benchmark datasets. We also find that the combination of FDS and SDS yields the best overall performance among all baselines. This is because by combining FDS and SDS, our method can utilize both the feature level information and the structural level information when comparing $\mathcal{L}_t$ and $\mathcal{U}_t$, hence we are able to effectively capture the influence of the labeled set on the value of unlabeled candidates. 

Although the EDS also combines both the structural and feature-level information of the nodes, it does not give as much performance gain as the other two scores. In Pubmed, we observe a performance drop when using EDS as our additional score. Moreover, we observe that combining all three dissimilarity scores are less powerful than combining FDS and SDS.  One possible explanation is that the embedding dissimilarity score is dependent on the quality of our GCN parameters. At the early stage of the training process, the GCN backbone might not be well trained, hence hard to provide an accurate embedding to aggregate the feature-level information and the structural information.

Another observation is that the GPF baseline performs especially well on the Cora dataset. One possible explanation is that Cora is a smooth dataset with a very high homophily ratio \cite{zhu2020beyond}. Therefore, the smoothness assumption in GPF formulation is properly met, making GPF very powerful.

\subsection{A closer look into FDS and SDS}

\begin{figure}[htbp]
\centering
\begin{minipage}[t]{0.48\textwidth}
\centering
\includegraphics[height=0.5\textwidth, trim = 25 200 20 200, clip]{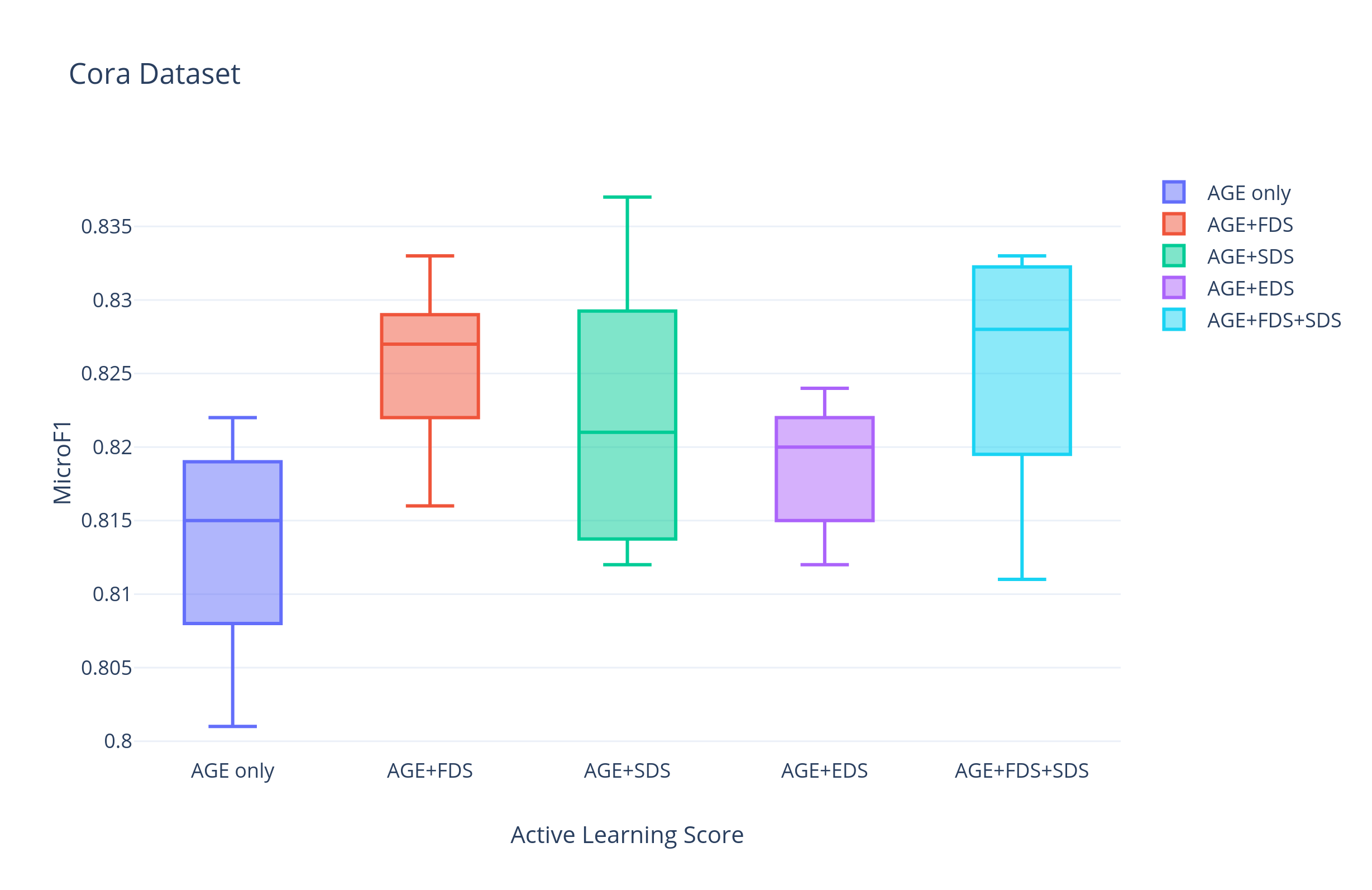}
\caption{Cora Micro F1}
\end{minipage}
\begin{minipage}[t]{0.48\textwidth}
\centering
\includegraphics[height=0.5\textwidth, trim = 25 200 20 200, clip]{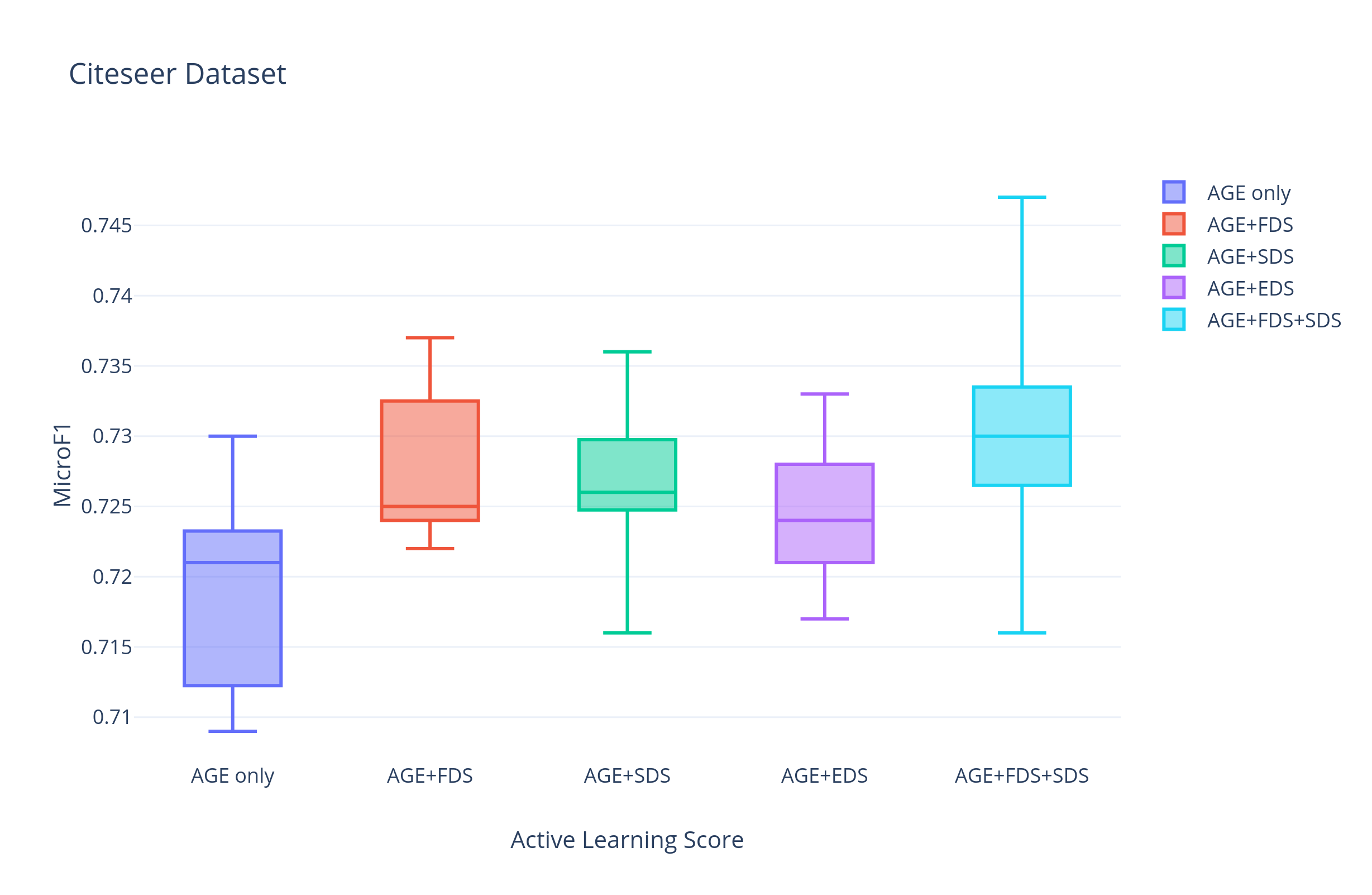}
\caption{Citeseer Micro F1}
\end{minipage}
\end{figure}

Figure 2 and Figure 3 provide us with a closer look at how stable the performance of different algorithms is. As we could see, AGE with FDS performs consistently better than AGE only, and the variance is relatively small. Compared with FDS, SDS is relatively more unstable. It could in some cases achieve superior performance than FDS, but perform worse in other cases (although better than the AGE baseline). It is likely those are the cases where the assumption in LINE \cite{tang2015line} breaks, i.e., two nodes are very different in their features although they share the same neighbors. We suspect that cases will be more frequent in heterophilic graph setting \cite{zhu2020beyond}, where two connected nodes are likely to be of different labels. We will discuss more active learning for heterophilic graph datasets in Appendix B.

\subsection{Ablation Studies}
We examine whether our method can be applied to GNN backbones other than GCN. As shown in Tables 2, 3, and 4, we replace the GCN backbone with graph attention network (GAT) \cite{velickovic2018graph} and simplifying graph convolutional network (SGC) \cite{pmlr-v97-wu19e}. We provide the results in Tables 2, 3, and 4. We observe that our three dissimilarity metrics could still improve the accuracy of node classification tasks when the network model is changed to GAT and SGC. This further assures the generalizability of our dissimilarity metrics among different GNN architectures.

\begin{table*}[h!]
\centering
\scriptsize
\begin{tabular}{cccccccc} 
 \toprule
 \textbf{\tiny{Backbone}} &
 \textbf{\tiny{Metrics}}& 
 \textbf{\tiny{Non-AL}}&
 \textbf{\tiny{AGE-only}} & 
 \textbf{\tiny{AGE+FDS}} & \textbf{\tiny{AGE+SDS}} & \textbf{\tiny{AGE+EDS}} & \textbf{\tiny{AGE+FDS+SDS}}\\
 \midrule
 
 \multirow{2}{*}{\textbf{\tiny{GAT}}}&\textbf{\tiny{MacroF1}} (\%) & 77.61 & 78.95 & 80.08 & 80.45 & 79.37 & \textbf{80.89}\\

&\textbf{\tiny{MicroF1}} (\%) & 78.28 & 80.39 & 80.93 & 81.76 & 80.46 & \textbf{82.32}\\

\midrule

 \multirow{2}{*}{\textbf{\tiny{SGC}}}&\textbf{\tiny{ MacroF1}} (\%) & 78.17 & 78.53 & \textbf{80.58} & 79.98 & 80.46 & 80.11\\

&\textbf{\tiny{MicroF1}} (\%) & 79.18 & 80.10 & \textbf{81.87} & 81.38 & 81.65 & 81.57\\

\bottomrule
\end{tabular}
\caption{F1 Scores of Different Active Learning Score Combination for Other Backbones for Cora}
\label{table:CORA}
\end{table*}

\begin{table*}[h!]
\centering
\scriptsize
\begin{tabular}{cccccccc} 
 \toprule
 \textbf{\tiny{Backbone}} &
 \textbf{\tiny{Metrics}}& 
 \textbf{\tiny{Non-AL}}&
 \textbf{\tiny{AGE-only}} & 
 \textbf{\tiny{AGE+FDS}} & \textbf{\tiny{AGE+SDS}} & \textbf{\tiny{AGE+EDS}} & \textbf{\tiny{AGE+FDS+SDS}}\\
 \midrule
 
 \multirow{2}{*}{\textbf{\tiny{GAT}}}&\textbf{\tiny{MacroF1}} (\%) & 62.05 & 61.96 & 61.90 & 63.22 & 61.95 & \textbf{63.10}\\

&\textbf{\tiny{MicroF1}} (\%) & 65.76 & 67.65 & 68.12 & 68.50 & 67.92 & \textbf{69.36}\\

\midrule

 \multirow{2}{*}{\textbf{\tiny{SGC}}}&\textbf{\tiny{ MacroF1}} (\%) & 64.61 & 65.94 & 66.01 & \textbf{67.25} & 66.94 & 66.14\\

&\textbf{\tiny{MicroF1}} (\%) & 68.25 & 70.65 & 70.24 & \textbf{71.72} & 71.12 & 70.63\\

\bottomrule
\end{tabular}
\caption{F1 Scores of Different Active Learning Score Combination for Other Backbones for Citeseer} 
\label{table:CITESEER}
\end{table*}

In addition, we examine whether our method can be applied to other benchmark datasets, such as heterophilic graphs whose connected nodes are less likely to belong to the same class \cite{zhu2020beyond}, and find that all the current scores will result in a degenerated performance. Our key assumption is that dissimilar nodes would be able to cover a more diverse training set (and can cover different classes), which is not true for heterophilic graphs. Details could be found in Appendix B.

\section{Conclusion and Future Directions}

In this work, with the aim of improving active learning on graphs, we propose 3 dissimilarity-based metrics to select the valuable nodes to query their labels. As shown by our experiments, all these scores are capable of bringing performance gain. In addition, our ablation study shows  the proposed metrics can work with different GNN backbones, which shows they have good generalizability. Though our proposed scores achieve a promising result, there are some limitations that we can work on as a future direction to further improve this work. First, since the EDS relies on the embedding learned by the backbone GNN model, its power might be weakened as the model may not be able to provide high-quality embeddings. Second, we mainly evaluate our method on homogeneous graphs now, and it is worth exploring to examine the method on more heterogeneous graphs.

\bibliographystyle{ACM-Reference-Format}
\bibliography{sample-base}

\newpage

\appendix

\section{Experimental Setting}

The Cora dataset consists of 2708 scientific publications that are classified into seven classes. And its graph has 5429 links. As for Citeseer, it has 3312 scientific publications classified into six classes. And its graph has 4732 links. On the other hand, we also have a larger dataset Pubmed which consists of 19717 publications that are classified into 3 categories. And it has 44338 links. For all the baselines, we re-run their experiments to make sure we get reliable results.

We choose 4 labels for each class to be our original labels. Then, we incrementally add 1 additional label after each epoch based on our new proposed scores. We use the same total label budget of $B$, which is 20 times the number of classes to better compare with our baseline. After the budget is reached, we no longer add more labels and continue training until converges.

Since the performance of active learning is often unstable, we run our experiments on 10 different validation sets using 10 different random seeds to reduce the variance of our results. We then take the average Macro F1 and Micro F1 over these 10 runs.

Here are some details of our hyper-parameter setting: We use the original implementation of GCN which has 2 graph convolution layers. We use a learning rate of 0.01 and a maximum number of epochs of 300. The hidden embedding dimension after layer 1 is 16. We choose our weight-tuning hyper-parameter $C$ to be 0.9 for Citeseer, 0.99 for Cora, and 0.995 for Pubmed. The hyper-parameter $C$ is chosen based on empirical results. One high-level explanation of our choice of $C$ is that Pubmed is denser than Cora, and Cora is denser than Citeseer. When the graphs are denser, we should put more weight on the centrality score, since the nodes with high centrality are more likely to be the most representation nodes in those graphs.

\section{A Closer Study on Heterophily}

In recent years, the study of heterophilic graph mining has gained some research attention. The homophily ratio stands for the fraction of graph edges where the two nodes they connect are of the same class. Zhu et. el. \cite{zhu2020beyond} have shown that algorithms performs well in homophilic graphs (e.g., Cora, Citeseer, Pubmed) might not perform as well in heterophilic datasets. Hence, it is worth checking the assumption of homophily in our active learning framework. In this way, we could better study the the generalizability of current AL framework in the setting of heterophilic graphs.

\begin{itemize}
\item \textbf{Centrality Score}: The effectiveness of the centrality score will somehow be affected by changing the assumption of homophily. While centrality usually implies representativeness on homophilic graphs, this might not be true on heterophilic graphs. In heterophilic graphs, all nodes with high centrality might belong to a single class. For example, in a university enrolment network with professors and students, all high centrality nodes are professors.
\item \textbf{Entropy Score}: The effectiveness of the entropy score will not be affected by changing the assumption of homophily. The reason is that the entropy score is generated using the model output, and it aims for evaluating how uncertain the model is when regarding particular nodes. Changing the graph from homophilic to heterophilic will not affect the pipeline which chooses the most uncertain nodes.
\item \textbf{Density Score and Embedding Dissimilarity Score}: The effectiveness of the density score will not be affected if we change the assumption of homophily. This is because most density scores are generated using clustering algorithms that take in model embedding outputs as raw inputs. If the model itself is well-trained, the quality of the node embedding will be good enough to provide guidance for valuable node selection. The same arguments also apply to EDS.
\item \textbf{Feature Dissimilarity Score}: The effectiveness of FDS will be severely affected by changing the assumption of homophily, since FDS is a score that completely neglects the structural information of a graph. While this is not a significant problem in homophilic graphs where nodes in different sub-graph clusters usually have different features, in heterophilic graphs, neglecting the structural information might lead to node selection within a single sub-graph cluster. This will severely affect the representativeness of the nodes being selected.
\item \textbf{Structural Dissimilarity Score}: The effectiveness of SDS will also be severely affected by changing the assumption of homophily, since the assumption "two nodes are similar if they share the same neighbors" will break under heterophilic setting.
\end{itemize}

To verify our hypothesises above, we conduct experiments on a heterophilic dataset, Chameleon, which contains 2277 nodes in 5 categories and 31421 edges. The data is collected by Rozemberczki et. al. \cite{musae} from the English Wikipedia which represents page-page networks on chameleons. Its homophily ratio is 0.23, which makes Chameleon a highly heterophilic dataset. For comparison, the homophily ratio is 0.81 in Cora, 0.74 in Citeseer, and 0.80 in Pubmed. We use the same experimental settings as what we have for the other experiments. For a fair comparison, we add only one score at a time when running on the Chameleon dataset. We use GCN-Cheby (GCC) to be our baseline, which is one of the SOTA models for node classification on heterophilic graphs \cite{zhu2020beyond}.

\begin{table}[h!]
\centering
\begin{tabular} {| c | c | c | c | c |} 
 \hline
\textbf{Score} &  \textbf{GCC} & \textbf{GCC+Centrality} & \textbf{GCC+Entropy} \\
 \hline 
 
 \textbf{MacroF1} (\%) & 47.68& 33.05 & 38.84 \\

\textbf{MicroF1} (\%) & 48.15& 33.70 & 40.33  \\
\hline
 
\end{tabular}

\begin{tabular} {| c | c | c | c | c |} 
 \hline
\textbf{Score} &  \textbf{GCC+FDS} & \textbf{GCC+SDS} & \textbf{GCC+EDS}\\
 \hline
 
 \textbf{MacroF1} (\%) &25.10 & 32.28 & 39.99  \\

\textbf{MicroF1} (\%) & 30.99& 32.58 & 40.41\\
\hline
 
\end{tabular}
\caption{F1 Scores of Different Dissimilarity Scoring Functions on Chameleon dataset}
\label{table:hetero}
\end{table}

The experimental results align with our hypothesis that FDS and SDS are the most severely affected scores when changed to the heterophilic setting. Those two scores have the most performance drop compared to GCC. The performance drop of the Centrality Score is also significant, but slightly better than FDS and SDS. In contrast, EDS and Entropy Score experience a relatively small performance drop compared to GCC, which means those two scores are more robust under a heterophilic setting. It is also notable that none of those scores performs better than GCC, which means that there is still room for performance improvement by finding specifically targeted types of active learning score functions for heterophilic graphs.

\end{document}